\documentclass[conference]{IEEEtran}
\IEEEoverridecommandlockouts
\usepackage{cite}
\usepackage{amsmath,amssymb,amsfonts}
\usepackage{algorithmic}
\usepackage{graphicx}
\usepackage{float}
\usepackage{textcomp}
\usepackage{xcolor}
\usepackage{multicol}
\usepackage{subcaption}
\usepackage{longtable}
\usepackage{lipsum}
\usepackage{booktabs, capt-of, tabularx}
\usepackage{cuted, stfloats}
\usepackage{hyperref}
\hypersetup{
    colorlinks=true,
    linkcolor=black,
    filecolor=black,
    citecolor=black,
    urlcolor=black,
}

\def\BibTeX{{\rm B\kern-.05em{\sc i\kern-.025em b}\kern-.08em
    T\kern-.1667em\lower.7ex\hbox{E}\kern-.125emX}}

\begin{document}

\title{Long Range Named Entity Recognition for Marathi Documents\\
}

\makeatletter
\newcommand{\newlineauthors}{%
  \end{@IEEEauthorhalign}\hfill\mbox{}\par
  \mbox{}\hfill\begin{@IEEEauthorhalign}
}
\makeatother

\author{\IEEEauthorblockN{Pranita Deshmukh}
\IEEEauthorblockA{
\textit{Pune Institute of Computer Technology}\\
\textit{L3Cube Pune}\\
Pune, India \\
dpranita9158@gmail.com}
\and
\IEEEauthorblockN{Nikita Kulkarni}
\IEEEauthorblockA{
\textit{Pune Institute of Computer Technology}\\
\textit{L3Cube Pune}\\
Pune, India \\
nikitakulkarni0108@gmail.com}
\and
\IEEEauthorblockN{Sanhita Kulkarni}
\IEEEauthorblockA{
\textit{Pune Institute of Computer Technology}\\
\textit{L3Cube Pune}\\
Pune, India \\
sanhitak17@gmail.com}
\and
\newlineauthors
\IEEEauthorblockN{Kareena Manghani}
\IEEEauthorblockA{
\textit{Pune Institute of Computer Technology}\\
\textit{L3Cube Pune}\\
Pune, India \\
kareenamanghani@gmail.com}
\and
\IEEEauthorblockN{Geetanjali Kale}
\IEEEauthorblockA{
\textit{Computer Department}\\
\textit{Pune Institute of Computer Technology}\\
Pune, India \\
hodcomp@pict.edu}
\and
\IEEEauthorblockN{Raviraj Joshi}
\IEEEauthorblockA{\textit{Indian Institute of Technology Madras} \\
\textit{L3Cube Pune}\\
Pune, India \\
ravirajoshi@gmail.com}
}

\maketitle

\begin{abstract}
The demand for sophisticated natural language processing (NLP) methods, particularly Named Entity Recognition (NER), has increased due to the exponential growth of Marathi-language digital content. In particular, NER is essential for recognizing distant entities and for arranging and understanding unstructured Marathi text data. With an emphasis on managing long-range entities, this paper offers a comprehensive analysis of current NER techniques designed for Marathi documents. It dives into current practices and investigates the BERT transformer model's potential for long-range Marathi NER. Along with analyzing the effectiveness of earlier methods, the report draws comparisons between NER in English literature and suggests adaptation strategies for Marathi literature. The paper discusses the difficulties caused by Marathi's particular linguistic traits and contextual subtleties while acknowledging NER's critical role in NLP. To conclude, this project is a major step forward in improving Marathi NER techniques, with potential wider applications across a range of NLP tasks and domains.

\end{abstract}


\section{Introduction}

Long Range Named Entity Recognition (NER) for Marathi text is a sophisticated natural language processing task that entails identifying and categorizing entities such as organizations, events, locations, and people within Marathi language documents. Unlike traditional NER systems, which typically focus on identifying entities within individual sentences, long-range NER extends this capability to encompass entities that span multiple sentences or even entire documents. This broader scope is essential for capturing complex relationships and contextual dependencies that may extend beyond the boundaries of individual sentences.

Developing effective Long Range NER for Marathi text involves overcoming several challenges unique to the language, including its morphological complexity, syntactic structure, and the presence of a rich variety of named entities specific to Marathi culture and society. Additionally, Marathi documents often exhibit diverse writing styles, ranging from formal literature to colloquial speech, further complicating the task of entity recognition \cite{joshi2022l3cube_mahacorpus}.

The applications of Long Range NER in Marathi text are manifold and far-reaching. Document classification benefits from the ability to accurately identify and categorize entities, enabling efficient organization and retrieval of information. Information extraction tasks are enhanced by the precise identification of entities related to organizations, events, and people, facilitating knowledge extraction from large volumes of unstructured text. Content recommendation systems can provide more personalized suggestions to Marathi-speaking users by leveraging insights derived from long-range entity recognition. Chatbots and virtual assistants equipped with Long Range NER capabilities can offer more context-aware and accurate responses in Marathi, enriching user interactions.

Moreover, Long Range NER supports sentiment analysis efforts in Marathi text, enabling organizations to monitor social media sentiment, analyze brand reputation, and conduct market research effectively. In the realm of academic research, Long Range NER assists scholars in analyzing Marathi literature, historical texts, and cultural documents, contributing to linguistic preservation and cultural studies. Furthermore, Long Range NER plays a crucial role in the preservation of cultural heritage by facilitating the recognition and cataloging of cultural entities within Marathi texts, thereby aiding in the documentation and dissemination of Marathi language and culture.

To achieve robust and accurate Long Range NER for Marathi text, advanced natural language processing techniques, including deep learning models such Transformer-based architectures, are employed. These models are trained on annotated Marathi language corpora, with careful attention paid to addressing language-specific challenges and ensuring generalization across diverse text genres and domains.

In summary, Long Range Named Entity Recognition for Marathi text represents a vital area of research and development in natural language processing, with significant implications for a wide range of applications, from information retrieval and content recommendation to cultural preservation and linguistic analysis. By harnessing the power of advanced NER techniques tailored to the nuances of the Marathi language, we can unlock new opportunities for understanding, analyzing, and leveraging Marathi language data in various domains and industries.

The paper is structured as follows: Section \ref{sec:Related Works} includes information about related work in this domain. Section \ref{sec:Datasets} describes the datasets used in the study.The models that were used are described in the \ref{sec:Methodology}, along with the methodology used model training and the accuracies obtained. Section \ref{sec:Result} provides the details about the results, and Section \ref{sec:Conclusion} consists of conclusions based on the results.

\section{Related Works}\label{sec:Related Works}

Mhaske et al.\cite{mhaske-etal-2023-naamapadam} present the Naamapadam dataset, a substantial collection covering 11 Indian languages, designed for Named Entity Recognition (NER). Their approach involves fine-tuning the mBERT model using the Naamapadam training data and subsequently evaluating it on the Naamapadam test data. The paper highlights the significance of this dataset and introduces an IndicNER model achieving an impressive F1 score of over 80 for 7 out of 9 test languages. By providing open-source access to both the dataset and the model, the authors contribute significantly to NER research in Indian languages.

Srivastava et al.\cite{srivastava2011named} focus on enhancing Named Entity Recognition (NER) accuracy for Hindi using a hybrid approach. They introduce the IJCNLP-08 manually annotated Hindi Named Entity Corpus and propose a methodology that combines Conditional Random Field (CRF)-based machine learning with rule-based methods. Additionally, they present a novel deep learning-based system tailored for Indian languages, incorporating affix embeddings and morphological analysis. By leveraging both machine learning and rule-based techniques, the authors achieve improved entity recognition accuracy, addressing a crucial aspect of NER research.

Sharma et al.\cite{article} tackle the challenge of Named Entity Recognition (NER) for Hindi, a resource-scarce language, by employing a hybrid approach. They utilize a Bi-LSTM-CNN-CRF hybrid model, incorporating Word2Vec and GloVe models for word representations and character- and word-level embeddings. Their experimental results demonstrate the superiority of their approach over baseline methods, including recurrent neural networks and LSTM. By enhancing NER accuracy, particularly for languages like Hindi, the authors contribute to advancing natural language processing systems, with potential applications in IoT devices and beyond.

Dahanayaka et al.\cite{7083904} address the challenges in Sinhala Named Entity Recognition (NER) by leveraging data-driven techniques. They utilize Conditional Random Fields (CRF) and Maximum Entropy Model to tackle the absence of capitalization features in Sinhala text. Through their experimentation, they highlight the effectiveness of statistical modeling in improving NER accuracy. The paper emphasizes the importance of balanced native-language training corpora for further enhancement in NER performance, particularly in languages with specific linguistic characteristics like Sinhala.

Litake et al.\cite{litake2022mono} focus on Named Entity Recognition (NER) tasks in Marathi and Hindi languages. They leverage state-of-the-art language models such as BERT, mBERT, ALBERT, and RoBERTa. Their findings suggest that monolingual models based on mahaBERT exhibit superior performance for Marathi NER compared to multilingual counterparts. By exploring various language models and their effectiveness in NER tasks, the authors contribute to advancing NER research, particularly in multilingual settings.

Murthy et al.\cite{murthy2018improving} propose a multilingual learning method for low-resource language Named Entity Recognition (NER). They employ a combination of Bi-LSTM and CNN, sharing network layers to improve performance across multiple languages. Through their experimentation on diverse datasets, including FIRE 2014 NER Corpus and ILCI Bengali Tourism Corpus, they showcase enhanced results and suggest future directions for handling linguistic diversity in NER tasks.

Nayel et al.\cite{nayel2019improving} introduce the FROBES model, an extension of the IOBES model for Named Entity Recognition (NER). They utilize Bi-LSTM and CRF to overcome the challenges posed by uni-directional LSTM models, particularly for multi-word recognition. Their proposed model outperforms other multi-word models, showcasing its effectiveness for entities with lengths greater than two. By addressing the limitations of existing models, the authors contribute to advancing NER accuracy, particularly in handling multi-word entities.

Patil et al. \cite{patil2016issues} delve into methodologies for crafting a Named Entity Recognition (NER) system for Marathi, a language posing various challenges such as agglutination and spelling variations. They explore different machine learning algorithms including HMM, MEMM, CRF, SVN, Adaboost, Decision Trees, Boot-strapping, and Clustering to address these complexities. By proposing and evaluating various approaches, the authors provide insights into the design and development of effective Marathi NER systems, crucial for organizing unstructured natural language data.

Patil et al. \cite{patil2022l3cubemahaner} introduce the L3Cube-MahaNER dataset, a golden standard for Marathi Named Entity Recognition (NER) consisting of 25,000 manually annotated sentences. They employ various models including CNN, LSTM, mBERT, XML-RoBERTa, IndicBERT, and MahaBERT to achieve optimal results. Their findings highlight the effectiveness of biLSTM for both IOB and non-IOB notations, with MahaRoBERTa and MahaBERT excelling in IOB and non-IOB formats, respectively. By introducing this dataset and evaluating multiple models, the authors contribute to advancing Marathi NER research, facilitating further development in this domain

\section{Datasets}\label{sec:Datasets}

\begin{table}[!h]
    \centering
    
    \label{tab:Datasets}
    \begin{tabular}{|l|c|c|c|}
        \hline
        & \textbf{Training} & \textbf{Testing} & \textbf{Validation} \\
        \hline
        O & 171956 & 16262 & 12119 \\
        BNEM & 5824 & 523 & 404 \\
        BNEP & 4775 & 428 & 322 \\
        BNEL & 4461 & 407 & 293 \\
        BNEO & 2741 & 256 & 178 \\
        INEP & 2135 & 191 & 141 \\
        BNED & 1937 & 183 & 135 \\
        INEO & 1435 & 129 & 90 \\
        INEM & 1228 & 97 & 84 \\
        BED & 838 & 74 & 61 \\
        BNETI & 633 & 63 & 43 \\
        INED & 529 & 53 & 41 \\
        INEL & 488 & 40 & 36 \\
        IED & 165 & 18 & 14 \\
        INETI & 111 & 10 & 5 \\
        \hline
    \end{tabular}
    \caption{Dataset Train-Test-Validation Split}
\end{table}

 We utilized the MahaNER dataset as our primary resource. This publicly available dataset comprises manually annotated Marathi language words classified into categories such as Person, Organization, Location, Measure, and Date. Derived from the MahaCorpus, this dataset is meticulously curated to ensure accuracy and representativeness.

The MahaNER dataset is provided in two formats: IOB (Inside Outside Beginning) and non-IOB. The IOB format, which we adopted for our study, structures the data to indicate the inside, outside, and beginning of named entities, facilitating more precise NER modeling.

Our chosen subset of the MahaNER dataset contains a total of 25,000 samples, comprising 21,500 training samples, 2,000 test samples, and 1,500 validation samples. These samples are categorized into 15 labels: "O," "BNEM," "BNEP," "BNEL," "BNEO," "BNED," "INEP," "INEO," "INEM," "BED," "BNETI," "INED," "INEL," "IED," and "INETI." However, it's notable that the distribution of labels is imbalanced, with a notably higher number of words categorized as "O" compared to other labels.

Each entry in the MahaNER dataset consists of three columns: Words, Labels, and sentence ID. The Words column contains individual words from the sentences, while the Labels column specifies the category of each word. The sentence ID column identifies the sentence to which each word belongs.

To address the focus of our research on long-range NER, we curated additional datasets by concatenating sentences from the MahaNER dataset. These concatenated datasets, namely concat2, concat3, and concat-similar, contain longer sentences to train our model effectively.

\begin{itemize}
    \item \textit{Concat2 Dataset}: This dataset involves the concatenation of two sentences randomly selected from the MahaNER dataset. The labels are concatenated along with the sentences, and the resulting sentences are further split into three columns: Words, Labels, and Sentence ID.
    
    \item \textit{Concat3 Dataset}: Similarly, the Concat3 dataset entails the concatenation of three sentences randomly chosen from the MahaNER dataset. The labels are combined accordingly, and the resulting data is structured into three columns: Words, Labels, and Sentence ID.
    
    \item \textit{Concat-similar Dataset}: In this dataset, we concatenate four sentences, with one of the sentences repeated for similarity. Again, the labels are appended accordingly, and the resulting dataset is structured into three columns: Words, Labels, and Sentence ID.
\end{itemize}

This process of dataset concatenation is applied to both the training, testing, and validation datasets, providing diverse and extended samples for robust model training and evaluation.

\begin{figure}[!h]
    \centering
    \includegraphics[scale=0.3]{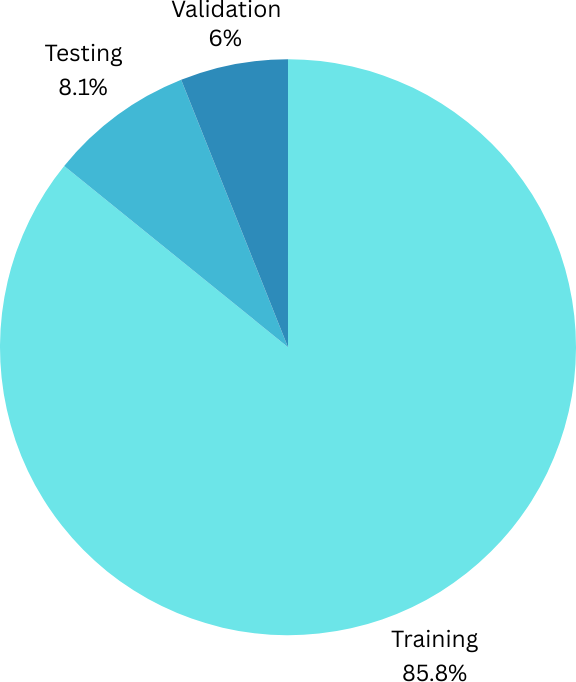}
    \caption{Train-Test Split}
    \label{fig:Datasets}
\end{figure}
\begin{table*}[h]
    \centering
    
    \label{tab:dataset_lengths}
    \begin{tabular}{|l|c|c|c|c|c|c|}
        \hline
        \textbf{Dataset} & \textbf{Training} & \textbf{Training} & \textbf{Testing} & \textbf{Testing} & \textbf{Validation} &\textbf{Validation}\\
        & \multicolumn{2}{c|}{\textbf{Min Length}} & \multicolumn{2}{c|}{\textbf{Min Length}} & \multicolumn{2}{c|}{\textbf{Min Length}} \\
        & \textbf{Min} & \textbf{Max} & \textbf{Min} & \textbf{Max} & \textbf{Min} & \textbf{Max} \\
        \hline
        Original (MahaNER) & 11 & 138 & 18 & 131 & 15 & 119 \\
        Concat 2 & 56 & 228 & 58 & 215 & 63 & 209 \\
        Concat 3 & 96 & 327 & 98 & 349 & 78 & 322 \\
        Concat Similar & 155 & 561 & 166 & 502 & 166 & 502 \\
        \hline
    \end{tabular}
    \caption{Dataset Lengths}
\end{table*}
\begin{figure}[!h]
    \centering
    \includegraphics[scale=0.32]{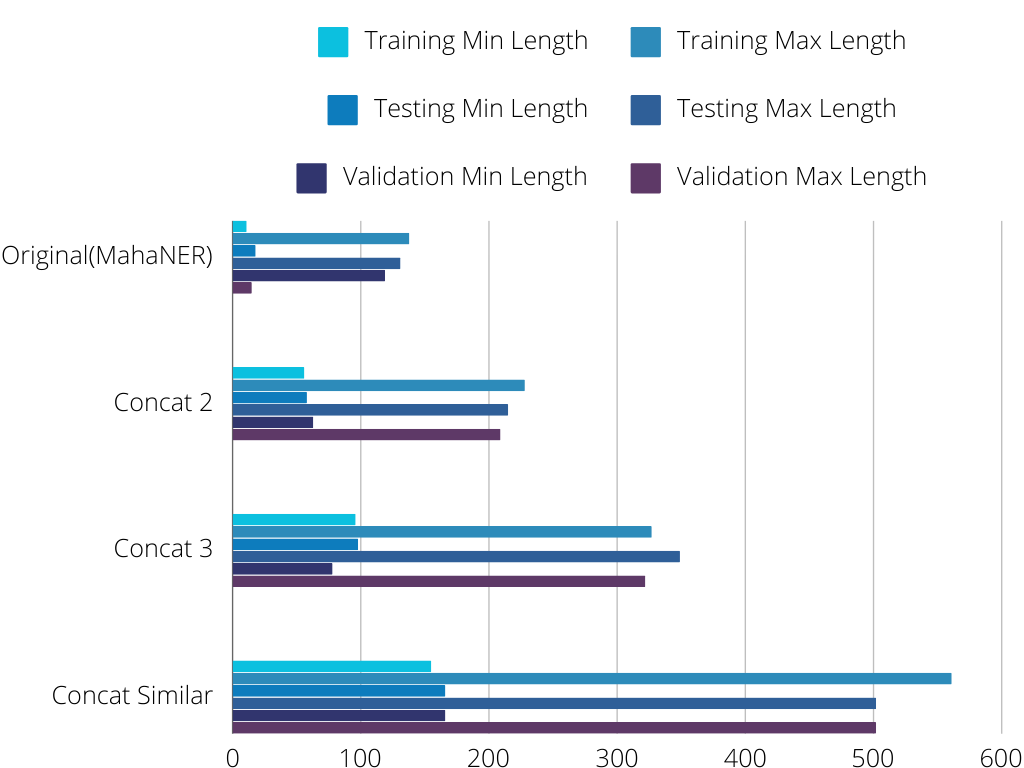}
    \caption{Sentence length Graph}
    \label{fig:Datasets}
\end{figure}

\newpage
\section{Methodology}\label{sec:Methodology}
The BERT based transformer models considered for this study is MahaBERT model.
\begin{figure*}[h]
    \centering
    \includegraphics[scale=0.45]{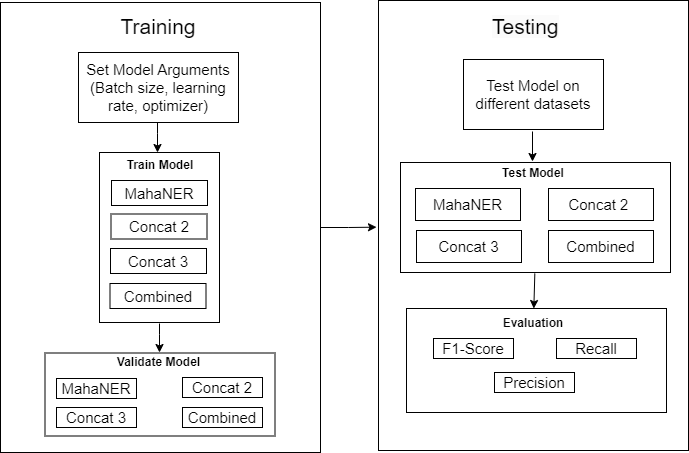}
    \caption{FlowChart}
    \label{fig:Datasets}
\end{figure*}
\subsection{Data Collection}
As mentioned above, we carefully constructed three diverse datasets - Concat2, Concat3, and Concat-similar - encompassing of various lengths of text for the purpose of long range. This ensured our model could generalize well across different contexts.
\subsection{Setting Model Arguments}
This preliminary phase entails the meticulous definition of critical hyperparameters governing the model's learning dynamics. These parameters, akin to finely tuned dials, intricately regulate the model's ability to distill insights from the provided data. Noteworthy among these hyperparameters are:

\begin{itemize}
    \item \textbf{Batch Size:} Dictating the quantity of data points presented to the model during each training iteration, the batch size profoundly influences the training dynamics. Smaller batches may yield meticulous optimization albeit at a slower pace, while larger batches might accelerate training albeit potentially at the expense of nuanced learning.
    
    \item \textbf{Learning Rate:} This pivotal parameter orchestrates the extent of adjustment made by the model's internal parameters in response to encountered errors. Striking a delicate balance, a higher learning rate expedites learning albeit risks instability and overfitting, while a lower rate engenders slower but potentially more stable convergence.
    
    \item \textbf{Optimizer:} Optimizers are algorithms used in machine learning to adjust the attributes of a model, such as weights and learning rate, to minimize the error or loss function during training. They play a crucial role in the training process, as they determine how quickly and effectively a model learns from the data. Common optimizers include Stochastic Gradient Descent (SGD) Adam, an adaptive optimization and many more. They can significantly impact the training process and the performance of the model, making it an important consideration in machine learning model development.
\end{itemize}

\subsection{Transformer Models}

We used the following model in our research:

\paragraph{MahaBERT}
MahaBERT\footnote{\url{https://huggingface.co/l3cube-pune/marathi-bert-v2}} is a Marathi BERT model based on Google's muril-base-cased version of the basic BERT architecture. It was fine-tuned using the L3Cube-MahaCorpus and several publicly available Marathi monolingual datasets. The model was initially pretrained using next sentence prediction (NSP) and masked language modeling (MLM) on 104 languages, including Marathi.
\subsection{Training the Model}
Herein lies the essence of model refinement. Armed with the defined hyperparameters, the model embarks on processing the training data, discerning intricate patterns and relationships within the features. Analogous to a scholar meticulously sifting through a vast repository of texts, the model gradually learns to discern and classify named entities, locations, or organizations within the data.
\subsection{Validation}
Validation plays a crucial role in assessing the performance of a trained model, ensuring that its predictions generalize well to unseen data. To prevent biased findings, a unique "test set" is reserved solely for performance evaluations, ensuring that the model is assessed on data it hasn't encountered during training. After the model has been trained, the test dataset is used to generate a variety of performance indicators, such as evaluation metrics, model outputs, and predictions. The validation process involves iterative execution, typically facilitated by a loop structure, allowing for multiple rounds of training and evaluation. This iterative approach enhances the comprehensiveness of performance assessment, enabling a more thorough evaluation of the model's predictive capabilities. For evaluation apart from the original dataset, we integrated three additional modified datasets: Concat2, Concat3, and Concat-similar, which we curated internally.

\subsection{Testing and Evaluation}
\subsubsection{Evaluation Metrics}
In consideration of the imbalanced nature of the dataset, we opted to utilize the F1 score as a primary evaluation metric. F1 score offers a balanced assessment, making it particularly suitable for imbalanced datasets where accurate classification of minority classes is crucial.Additionally, we employed precision and recall as supplementary evaluation metrics. Precision measures the accuracy of positive predictions, while recall assesses the model's ability to correctly identify all relevant instances of a class.
\subsubsection{Testing}
\begin{table}[!h]
    \centering
    \begin{tabular}{|c|c|c|c|c|}
        \hline
        \textbf{Train} & \textbf{Test} & \textbf{F1-Score} & \textbf{Precision} & \textbf{Recall} \\
        \hline
        Original & Original & 84.19 & 83.82 & 84.55 \\
        \hline
        Original &Concat 2 &76.64 &80.58 &73.06 \\
        \hline
        Original & Concat 3& 75.89&79.79 & 72.35\\
        \hline
        Original & Concat-similar&72.64 &75.8 &70.2 \\
        \hline
    \end{tabular}
    \caption{Original Dataset F1-Score Values.}
    \label{tab:Table I}
\end{table}
\begin{figure}[h]
    \centering
    \includegraphics[scale=0.20]{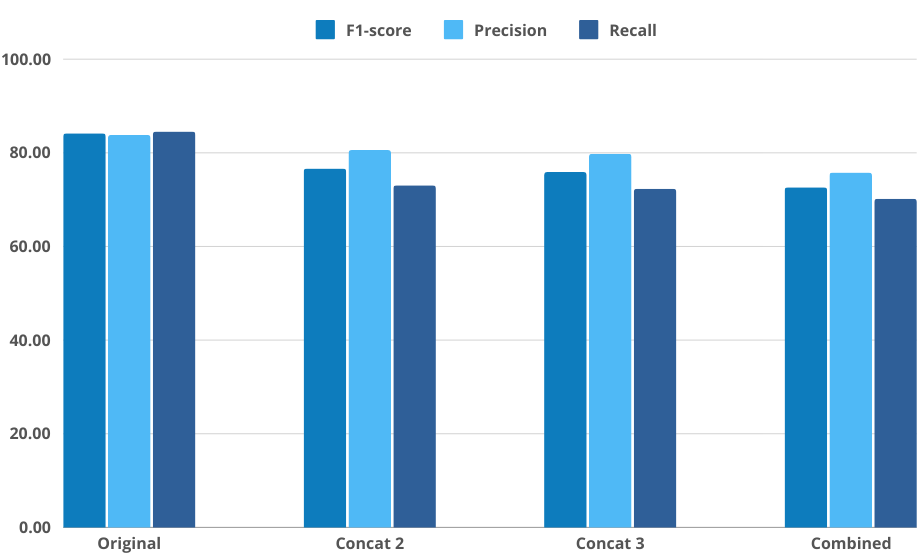}
    \caption{Original}
    \label{fig:Graphs}
\end{figure}

\begin{table}[!h]
    \centering
    \begin{tabular}{|c|c|c|c|c|}
        \hline
        \textbf{Train} & \textbf{Test} & \textbf{F1-Score} & \textbf{Precision} & \textbf{Recall} \\
        \hline
        Concat 2 & Original & 84.12 & 83.19 & 85.07 \\
        \hline
        Concat 2 &Concat 2 &83.73 &82.79 &84.7 \\
        \hline
        Concat 2 & Concat 3& 83.19&82.11 &84.29\\
        \hline
        Concat 2 & Concat-similar&82.4 &83.1 &83.5 \\
        \hline
    \end{tabular}
    \caption{Concat2 Dataset F1-Score Values}
    \label{tab:Table II}
\end{table}
\begin{figure}[!h]
    \centering
    \includegraphics[scale=0.20]{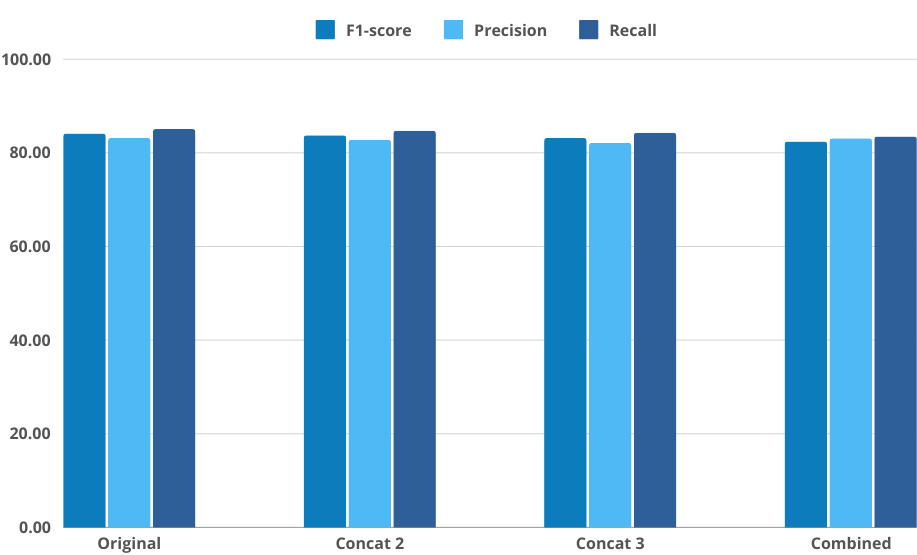}
    \caption{Concat2}
    \label{fig:Graphs}
\end{figure}
\begin{table}[!h]
    \centering
    \begin{tabular}{|c|c|c|c|c|}
        \hline
        \textbf{Train} & \textbf{Test} & \textbf{F1-Score} & \textbf{Precision} & \textbf{Recall} \\
        \hline
        Concat 3 & Original & 83.74&	83.29	&84.19 \\
        \hline
        Concat 3 &Concat 2 &83.97&	82.67	&85.31 \\
        \hline
        Concat 3 & Concat 3& 83.179&	82.206&	84.175\\
        \hline
        Concat 3 & Concat-similar&82.1&	83.94&	82.84 \\
        \hline
    \end{tabular}
    \caption{Concat3 Dataset F1-Score Values}
    \label{tab:Table III}
\end{table}
\begin{figure}[!h]
    \centering
    \includegraphics[scale=0.20]{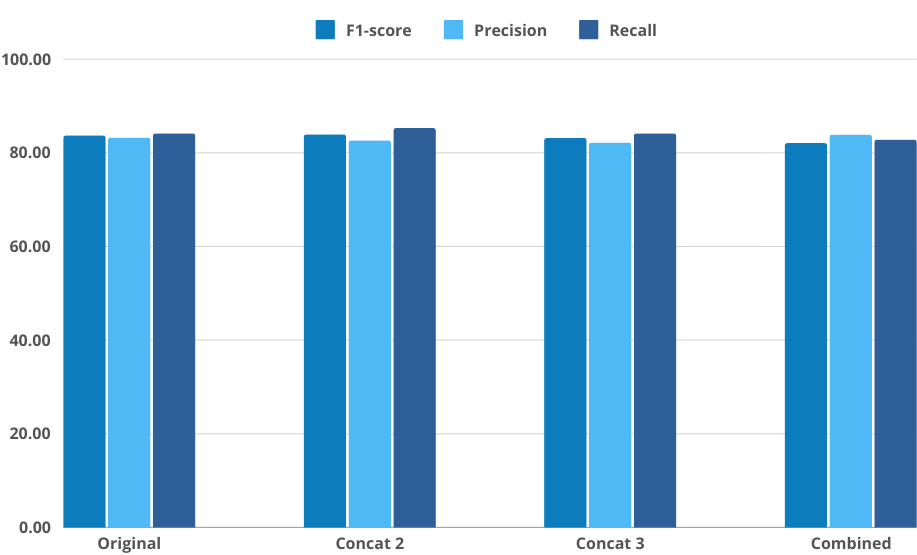}
    \caption{Concat3}
    \label{fig:Graphs}
\end{figure}
\begin{table}[!h]
    \centering
    \begin{tabular}{|c|c|c|c|c|}
        \hline
        \textbf{Train} & \textbf{Test} & \textbf{F1-Score} & \textbf{Precision} & \textbf{Recall} \\
        \hline
        Combined & Original & 83.14&	83.78	&83.46 \\
        \hline
        Combined &Concat 2 &83.89&	84.9	&82.91 \\
        \hline
        Combined & Concat 3& 82.61	&81.5	&83.75\\
        \hline
        Combined & Concat-similar&85.46&	84.14	&84.78 \\
        \hline
    \end{tabular}
    \caption{Concat-similar Dataset F1-Score Values}
    \label{tab:Table IV}
\end{table}
\begin{figure}[!h]
    \centering
    \includegraphics[scale=0.20]{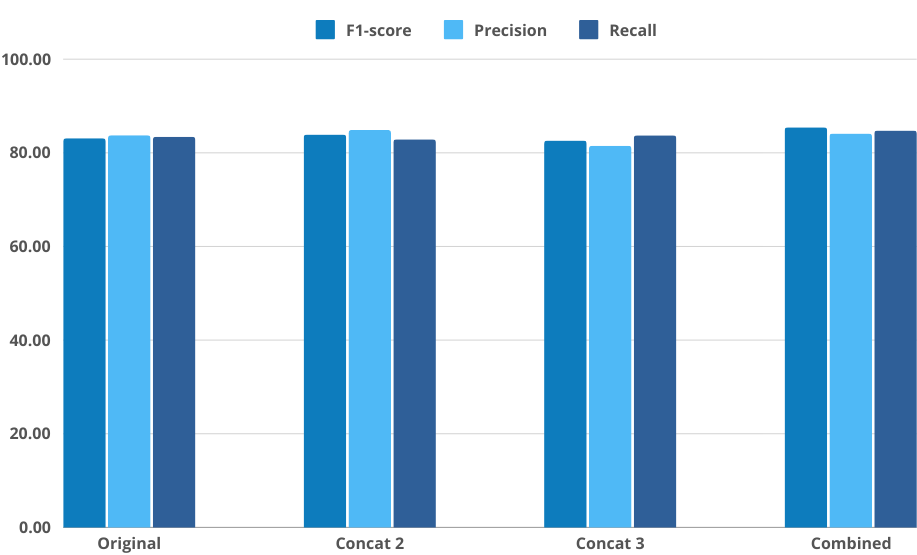}
    \caption{Concat-similar}
    \label{fig:Graphs}
\end{figure}

\section{Result}
\label{sec:Result}
Tables \ref{tab:Table II}, \ref{tab:Table III}, \ref{tab:Table IV} and \ref{tab:Table IV} show the results of comparison of the evaluation metrics when Original, Concat2, Concat3 and Concat-similar datasets are used as train data respectively. These datasets were provided as train data to the fine-tuned MahaBERT model and were tested on the datasets having both smaller and larger range texts as compared to the train data. 


As seen in the above tables it can be inferred that when the testing dataset has long range texts the F1-score improves but it can also be inferred that when the range of the testing dataset is larger or comparable to the training data it retrieves comparative as well. 

It can be observed that when the original dataset is taken as the training data it does not provide desired result on long range text testing data. In case of Concat2 as the training data comparative F1-scores are obtained with original, Concat2, Concat3 and Concat-similar datasets as the test data with Original as the test data retrieving the highest value of 84.12. When Concat3 is used as the training data the same is observed with highest F1-Score of 83.97 when Concat2 is used as the testing data. While for Combined dataset as the training data highest F1-Score value of 85.46 is obtained when the Concat-similar dataset is used as the testing data.

Overall it was observed that as the range of text in the training data sentences is increased a comparative improvement is obtained and eventually the desired results are obtained.



\section{Conclusion}\label{sec:Conclusion}
In this study we explore Marathi and how Long Range Named Entity Recognition can be employed for it which is one of the most NLP techniques. This study offers a thorough examination of MahaBERT tranformer model on self-curated and preprocessed Original, Concat2, Concat3 and Concat-similar datasets respectively.

Furthermore, this paper illustrates how the MahaBERT model was fine-tuned and trained and tested on the above mentioned datasets.  This study also solves the problem of Named Entity Recognition on Long Range texts and also yields desired outcomes

Altogether this study provides insights about how tranformer models can be employed for Long Range NER and also lays groundwork for future improvements low resource language and how various NLP techniques such as NER can be employed on them. As the environment of digital information continues to shift, researchers and practitioners seeking to enhance NLP applications in diverse linguistic contexts might gain from the methods and conclusions presented in this work.
\section*{Future Scope}
To further enhance the capabilities and impact of this project in long-range Named Entity Recognition (NER) for Marathi, several key areas can be targeted for future work. Further enhancement can be achieved by training the model using additional datasets from diverse domains, such as legal documents and social media, to improve its adaptability. Also, the application of the model can be extended to other low-resource languages, enhancing NER capabilities in underrepresented linguistic communities. Integrating the model with various NLP applications can enhance its usability and effectiveness in real-world scenarios. Lastly, exploring alternative deep learning techniques can lead to improvements in model efficiency and performance. These advancements have the potential to significantly enhance the model's impact and utility in the field of natural language processing.
\section*{Acknowledgements} 
This work was completed as part of the L3Cube Mentorship Program in Pune. We would like to convey our thankfulness to our L3Cube mentors for their ongoing support and inspiration. This work is a part of L3Cube-MahaNLP initiative \cite{joshi2022l3cube_mahanlp}.

\bibliography{main}

\begin{thebibliography}{10}

\bibitem{joshi2022l3cube_mahacorpus}
Raviraj Joshi.
\newblock L3cube-mahacorpus and mahabert: Marathi monolingual corpus, marathi bert language models, and resources.
\newblock In {\em Proceedings of the WILDRE-6 Workshop within the 13th Language Resources and Evaluation Conference}, pages 97--101, 2022.

\bibitem{mhaske-etal-2023-naamapadam}
Arnav Mhaske, Harshit Kedia, Sumanth Doddapaneni, Mitesh~M. Khapra, Pratyush Kumar, Rudra Murthy, and Anoop Kunchukuttan.
\newblock Naamapadam: A large-scale named entity annotated data for {I}ndic languages.
\newblock In Anna Rogers, Jordan Boyd-Graber, and Naoaki Okazaki, editors, {\em Proceedings of the 61st Annual Meeting of the Association for Computational Linguistics (Volume 1: Long Papers)}, pages 10441--10456, Toronto, Canada, July 2023. Association for Computational Linguistics.

\bibitem{srivastava2011named}
Shilpi Srivastava, Mukund Sanglikar, and DC~Kothari.
\newblock Named entity recognition system for hindi language: a hybrid approach.
\newblock {\em International Journal of Computational Linguistics (IJCL)}, 2(1):10--23, 2011.

\bibitem{article}
Richa Sharma, Sudha Morwal, Basant Agarwal, Ramesh Chandra, and Mohammad~Shoeb Khan.
\newblock A deep neural network-based model for named entity recognition for hindi language.
\newblock {\em Neural Computing and Applications}, 32, 10 2020.

\bibitem{7083904}
J.~K. Dahanayaka and A.~R. Weerasinghe.
\newblock Named entity recognition for sinhala language.
\newblock In {\em 2014 14th International Conference on Advances in ICT for Emerging Regions (ICTer)}, pages 215--220, 2014.

\bibitem{litake2022mono}
Onkar Litake, Maithili Sabane, Parth Patil, Aparna Ranade, and Raviraj Joshi.
\newblock Mono vs multilingual bert: A case study in hindi and marathi named entity recognition.
\newblock {\em arXiv preprint arXiv:2203.12907}, 2022.

\bibitem{murthy2018improving}
Rudra Murthy, Mitesh~M Khapra, and Pushpak Bhattacharyya.
\newblock Improving ner tagging performance in low-resource languages via multilingual learning.
\newblock {\em ACM Transactions on Asian and Low-Resource Language Information Processing (TALLIP)}, 18(2):1--20, 2018.

\bibitem{nayel2019improving}
Hamada~A Nayel, HL~Shashirekha, Hiroyuki Shindo, and Yuji Matsumoto.
\newblock Improving multi-word entity recognition for biomedical texts.
\newblock {\em arXiv preprint arXiv:1908.05691}, 2019.

\bibitem{patil2016issues}
Nita Patil, Ajay~S Patil, and BV~Pawar.
\newblock Issues and challenges in marathi named entity recognition.
\newblock {\em International Journal on Natural Language Computing (IJNLC)}, 5(1):15--30, 2016.

\bibitem{patil2022l3cubemahaner}
Parth Patil, Aparna Ranade, Maithili Sabane, Onkar Litake, and Raviraj Joshi.
\newblock L3cube-mahaner: A marathi named entity recognition dataset and bert models, 2022.

\bibitem{joshi2022l3cube_mahanlp}
Raviraj Joshi.
\newblock L3cube-mahanlp: Marathi natural language processing datasets, models, and library.
\newblock {\em arXiv preprint arXiv:2205.14728}, 2022.

\end{thebibliography}
\bibliographystyle{unsrt}
\end{document}